# Investigating the relationship between empathy and attribution of mental states to robots

Alberto Lillo[1], Alessandro Saracco[1], Elena Siletto[1], Claudio Mattutino[1] and Cristina Gena[1]

[1]*Department of Computer Science, University of Turin, Italy*

**Abstract**

This paper describes an experimental evaluation aimed at detecting the users' perception of the robot's empathic abilities during a conversation. The results have been then analyzed to search for a possible relationship between the perceived empathy and the attribution of mental states to the robot, namely the user's perception of the robot's mental qualities as compared to humans. The involved sample consisted of 68 subjects, including 34 adults and 34 between teenagers and children. By conducting the experiment with both adult and child participants, make possible to compare the results obtained from each group and identify any differences in perception between the various age groups.

## 1. Introduction

The advent of Artificial Intelligence and robotic technologies has ushered in an era of extraordinary potential for humanity, radically transforming the way we live, work, and interact. Within this context, Human-Robot Interaction (HRI) [1] emerges as a critically important field of study, at the crossroads of technological innovation and humanistic understanding. As technological progress leaps forward, crucial issues arise not only about how robots can assist humans in their daily activities, such as work [2], school [3], home [4], cleaning and caring for vulnerable persons [5], but also about how they can harmoniously integrate into the social dynamics that characterise our existence. Thus, the primary challenge of HRI is not merely technical but also profoundly relational: how to design robots that are not perceived as mere machines, but as social companions, capable of empathy and meaningful interaction with human beings [6, 7], supporting them in their daily lives and their preferential choices [8].

This question opens the way for a broader reflection on the meaning of *empathy* in the robotic domain [9, 10] and the role it can play in facilitating an effective and positive social integration of robots. Empathy, traditionally understood as the ability to comprehend and share the feelings of others, becomes a desirable quality for robots as well, especially in areas where the human-robot relationship is crucial, such as elderly care, education, and therapeutic support. The design of empathic robots, however, raises complex questions, not only of a technological nature but also philosophical and ethical: is it possible for a machine to possess a true understanding of human emotions? And how does robotic empathy impact human perception of these new social entities?

Simultaneously, advancements in HRI research highlight the importance of developing increasingly intuitive and natural human-robot interfaces [11, 12, 13], enabling fluid and bidirectional communication. In this sense, robot design cannot overlook a holistic approach that considers not only functional aspects but also emotional and relational ones, designing machines capable of "understanding" and adapting to the human context in which they are inserted.

## 2. State of the art

Studies have shown that robots that can adjust their behaviour according to the affective state or personality of a user are more accepted as interaction partners [14] and are seen as more friendly, caring, sympathetic, supportive and trustworthy. Therefore, several empathic models for social robots have been proposed [15, 16]. Succeeding in this daunting challenge can have profound positive effects on users' attitudes towards social robots. Responding to the user's affective experience in a socially appropriate manner is considered crucial to achieving user trust and satisfaction. In one experiment it was found that the ability of a robot to respond with its empathic system in a situationally appropriate manner is more important for comforting the user than a sophisticated and detailed recognition of affect [17].

However, reacting empathically requires the robot's recognition of the user's emotional state. This knowledge is challenging as it requires an evaluation of a deeply personal and individual experience and for these reasons errors are likely to occur. This reinforces the importance of understanding how people respond to empathic capabilities if a robot behaves incongruently with the user's emotional experience. Inaccurate emotional responses may indeed have negative consequences on users' evaluations of an agent. Furthermore, virtual agents that display emotions incongruent to the situation are also less appreciated by users than those that do not express any emotion at all [18].

Research on the subject, however, has yet to fully uncover the effects of empathic behaviour in different situations, including possible inaccurate responses [18]. According to several neurological and psychological researches ([19, 20] the involvement of the mirror neuron system is implicated in neurocognitive functions, such as social cognition, language, empathy and the Theory of Mind (ToM) [21, 22], which is a human-specific ability to attribute mental states - intentions, thoughts, desires and emotions - to oneself and others to explain and predict behaviour. Specifically, attribution of mental states (AMS) has been defined as "*the cognitive ability to reflect on one's own and others' mental states, such as beliefs, desires, feelings, and intentions*" [23]. In [24], the authors presented an experimental study showing that the humanoid robot NAO is able to stimulate the attribution of mental states towards itself when it stimulate empathy. This result suggests a possible correlation between empathy toward the robot and humans' attribution of mental states to it.

# 3. The experiment

In this Section we introduce an experimental evaluation aimed at detecting the users' perception of the robot's empathic abilities during a conversation. The results have been then analyzed to search for a possible relationship between the perceived emapthy and the attribution of mental states to the robot, namely the user's perception of the robot's mental qualities as compared to humans. The involved sample consisted of 68 subjects, including 34 adults and 34 between teenagers and children. By conducting the experiment with both adult and child participants, make possible to compare the results obtained from each group and identify any differences in perception between the various age groups.

**Methodology and experimental design.** To achieve the most significant results, it's crucial to adhere to a clear and well-defined methodology. The experiment's organization was meticulously planned to ensure its effective and efficient execution, leading to valid and reliable outcomes. The planning process began with identifying the user categories participating in the experiment. The 'adults' category includes users aged 18 and above, while the 'children and young people' category comprises users aged 17 and below.

Each user category, the adults and the youth and children, will be split into two groups: the control group and the experimental group. The control group won't be subjected to any changes in the independent variable, serving as a key reference point. This setup allows for the evaluation of the manipulation's impact by comparing the results of the two groups.

**Experimental group.** Users in this group will interact with an expressive robot that can respond to the user's emotions and express its own state. The robot will also make movements during the conversation to facilitate non-verbal interaction.

**Control group.** Users in the control group will interact with an apathetic robot, which is programmed to complete tasks without showing empathy towards the user's emotions. The robot will exhibit a less enthusiastic and more static demeanor, with no specific movements to aid non-verbal interaction.

**Independent variable.** The independent variable is the social and emotional skill level that will be implemented in the NAO virtual robot. Specifically, there are two conditions for the conduct of the experiment: i) Emotional and empathic robot; ii)n Apathetic robot. In the course of the experiment, this variable will be manipulated in order to test the interaction with the two types of robots and to record the differences in users' perceptions.

**Dependent variables.** The primary dependent variable is the users' reactions to the different experimental conditions, encompassing all measurements and observations of users' responses post-interaction with the robot. These responses will be primarily gauged through a structured interview. Another potential dependent variable is task performance, with a focus on whether the user successfully completes the intended task. External variables that could impact the results should also be considered. Analyzing the dependent variables' data will enable the assessment of the independent variable manipulation's effect on the research.

**Sample Selection.** When conducting experiments, it's crucial to choose a representative sample to prevent bias. However, for this experiment, the sample wasn't randomly selected but was chosen from a readily available and willing group. Particularly, most children in the sample were from a cooperative dance school. Therefore, while the sample isn't fully representative of all user categories, it provides a solid foundation for future research.

**Measurement and Instruments.** The data collection for the dependent variables will utilize quantitative measurements, which allows for numerical data collection and subsequent statistical analysis. Initially, a questionnaire was selected as the most appropriate method for collecting this data. However, considering the administration method, a structured interview was introduced instead. This method, unlike questionnaires, provides the opportunity to clarify questions, aiding participants, including children, to fully understand and comfortably participate in data collection. The interview process involves asking the user to score each question on a rating scale. Specific tools have been utilized for this process. *Batson's self-assessment*: This assessment [25] asks participants to rate their experience of specific emotions on a scale from 1 to 5. For the Batson self-assessment, users will score their emotions on a scale from 'Not at all' to 'Totally'. The assessment will evaluate 23 emotions, expressed through the following adjectives: frightened, suffering, sympathetic, sensitive, agitated, cordial, worried, stressed, sad, compassionate, upset, tender, distressed, impressed, downhearted, depressed, afflicted, annoyed, kind, melancholic, moved, and uncomfortable.

*AMS-Q questionnaire*: the administration of this questionnaire [26, 27] will allow us to perceive the degree to which users attribute mental states to the NAO robot. The test consists of 25 questions and asks users to rate whether they think the robot (e.g. "can you understand?", "can it decide?", "can you tell a lie?", "can you try to do something?").

Upon completion of the experiment and data collection, various quantitative analyses will be conducted using Excel. For the Batson self-assessment results [25], the mean score and standard deviation for each emotion will be calculated for each group. A T-test will then be performed to determine if there's a significant difference between the means of the experimental and control groups. The AMS questionnaire [26, 27] results will undergo a similar analysis, but will first be divided into five mental state categories. T-tests will be calculated for both the experimental and control groups as a whole, and separately for children and adults. This allows for data interpretation across both groups and distinct age categories.

**Experimental plan**. Each participant, numbered 1 to 34 based on their participation order, will be randomly assigned to either the experimental or control group. A random number generator was used for this assignment, with the first 17 numbers allocated to the experimental group and the remaining 17 to the control group. This procedure applies to both 'adults' and 'children and young people' categories. The test will proceed in multiple stages. Each of these steps has a precise objective and are designed to be able to collect valid and reliable data:

*Introduction*. In this initial stage, the user will be greeted and given a brief introduction. Only essential details for interacting with NAO will be provided at this time, while answers to additional inquiries will be deferred until the experiment's conclusion.

*Interaction and task.* In this stage, the user will interact with the NAO virtual robot. The session will start with an introduction between the robot and the user. Following topics will include interests and family, culminating in the final task.

*Gathering quantitative information.* Once the test has been completed, the user will immediately be given a questionnaire containing the necessary quantitative measurements, namely, the Batson self-assessment [25] and the AMS questionnaire [26, 27].

*Closing.* In this last phase we will move on to the final greetings and thanks, answering users' questions and curiosities.

## 4. Creating the personality of the virtual robot NAO

The process of defining the personality of NAO constituted the first step necessary for the implementation of the robot. For this implementation, we focused on the creation of a Personas. This step then made it possible to program the robot and write dialogues that were consistent with each other and with the robot's personality.

In order to describe and frame the desired personality of the robot, the Big Five [28] test was performed. Thanks to this test, it was possible to optimise and think about character traits. In a first step, the job that NAO could hypothetically perform was chosen. The professional figure identified was the teacher. This, it was assessed, would fit well with the envisaged personality and the subsequent task.

During the following phase, the main characteristics to be attributed to the robot were chosen, which were considered fundamental, such as the traits: calm, patient and wise. Once these elements had been identified, the test was completed. The results obtained as follows.

*Neuroticism: 37 out of 120.* The robot was chosen to have a very positive attitude that rarely experiences negative emotions. He is characterised by a sunny and patient manner. These traits were also favoured with a view to safe interaction.

*Extroversion: 88 out of 120.* As discussed extensively in chapter one, robots need to be able to communicate and engage in interaction with ease in order to be recognised as social agents. This character trait emerged from this reflection. In particular, high scores in this category describe a sociable and assertive personality. This could lead NAO to make friends very quickly and relate to users.

*Openness to experience: 88 out of 120.* In order to model a robot that would then be credible when tested with users, it was decided not to attribute characteristics to NAO that could be considered unthinkable. Its character was, therefore, calibrated to be based on facts.

*Conscientiousness: 106 out of 120.* This is the category where NAO scored the highest, since it has been designed to be a responsible, organised and disciplined robot. Furthermore, questions concerning levels of confidence in one's own abilities were always answered as 'agree' and 'very agree'.

*Pleasantness: 101 out of 120.* In terms of values, sincerity and the spirit of cooperation were favored. Those who score high on this trait are also characterized by kindness and altruism.

## 5. Results and Comparisons

Comparing the Batson self-assessment results, both the experimental and control groups reported minimal negative emotions, with over 88% stating they felt 'Not at all' distressed, worried, stressed, sad, upset, downhearted, depressed, distressed, and annoyed. However, differences emerged in positive emotions, with 100% of the experimental group feeling 'Totally' nice and kind, compared to 76.5% in the control group. The T-test showed minimal significance levels for 'Sympathetic' (T-test= 0.04) and 'Kind' (T-test= 0.047). Regarding the AMS questionnaire, similar average scores were observed in all dimensions for both groups: epistemic (7.5 vs. 8.9), emotional (7.5 vs. 7.7), desires and intentions (8.8 vs. 9.7), imagination (7.6 vs. 8.8), and perceptual (6.8 vs. 6.7). The T-test did not reveal any significance, but it's noteworthy that high average scores were obtained in each dimension for both groups.

Comparing the Batson Self-Assessment results, both the experimental and control groups reported minimal negative emotions, with over 82% stating they felt 'Not at all' scared, hurt, sad, upset, distressed, downhearted, depressed, annoyed, melancholic, and uncomfortable. However, differences emerged in positive emotions. The experimental group reported total sympathy for the robot by 64.75%, while the control group reported 41.25The experimental group, which interacted with a more emotional robot, reported feeling somewhat more sensitive, with only 23.5% saying 'Not at all', compared to 53% in the control group. This difference was significant (T-test=0.03). The experimental group also reported feeling more friendly towards the robot, with 82.4% reporting 'Totally', compared to only 29.5% in the control group. This emotion was found to be significant (T-test=0.01). For the feeling 'impressed', no subjects in the experimental group reported 'Not at all' and 41.25% indicated 'Very much'. In contrast, 64.7% of the subjects in the control group answered 'Not at all' and none indicated 'Totally'. This emotion was also found to be significant (T-test=0.000001). Regarding the AMS questionnaire, divergent averages were noted for all dimensions between the control and experimental group: epistemic (5.5 vs. 8.2), emotional (1.9 vs. 7.5), desires and intentions (4.0 vs. 7.9), imagination (2.0 vs. 6.9), and perceptual (2.4 vs. 5.6). Upon applying the T-test, the dimensions epistemic (T-test=0.01), emotional (T-test=0.000001), desires and intentions (T-test=0.001), imagination (T-test=0.000001), and perceptual (T-test=0.0003) were found to be significant.

Comparing the results between the two categories, children and adults, from the Batson's test showed that both groups empathized with the robot, with higher averages for positive emotions. However, adults in the experimental group reported higher averages for positive emotions than the control group. In the AMS questionnaire, no clear differences emerged between the experimental and control groups for children. However, adults in the experimental group attributed higher scores to the robot in all dimensions. For the epistemic dimension, children in the experimental group had a mean of 8.9 and adults 8.2, while the control group had 7.5 for children and 5.5 for adults. For the emotional dimension, the experimental group averaged 7.7 for children and 7.5 for adults, while the control group had 7.5 for children and 1.9 for adults. For the dimension of desires and intentions, the experimental group averaged 9.7 for children and 7.9 for adults, while the control group had 8.8 for children and 4 for adults. For the imagination dimension, the experimental group had a mean of 8.8 for children and 6.9 for adults, while the control group had 7.6 for children and 2 for adults. Finally, for the perceptual dimension, the experimental group averaged 6.7 for children and 5.6 for adults, while the control group had 6.8 for children and 2.4 for adults.

## 6. Discussion and Conclusion

The paper examined the relationship between Human-Robot Interaction (HRI) and empathy, and the role of social robots. It used a semi-structured interview method for data collection, which was effective in maintaining user concentration. The study found that the personal background of users, especially adults, significantly influenced their responses. The results revealed that adults in the experimental group attributed more mental states to the robot, suggesting that the robot's additional features, such as movements and positive mood, made it appear more capable.

However, these features had no impact on younger users, who were consistently enthusiastic and perceived the robot as almost limitless. In conclusion, the paper highlighted the importance of emotions and empathy in HRI and suggested that these factors will be crucial in the future development of social robots.